\documentclass[letterpaper, 10 pt, conference]{ieeeconf}  

\IEEEoverridecommandlockouts                              

\usepackage{amsmath}
\usepackage{xcolor, soul}
\usepackage{subcaption}
\usepackage{amsmath}
\usepackage{amsfonts}
\usepackage{amssymb}
\usepackage{bm}
\usepackage{multicol}
\usepackage{multirow}
\usepackage{optidef}
\usepackage{scalerel}
\usepackage{tikz}
\usetikzlibrary{svg.path}
\usepackage{colortbl}
\usepackage{color}
\definecolor{celadon}{rgb}{0.67, 0.88, 0.69}
\usepackage{algorithm}
\usepackage{algpseudocode}
\usepackage{graphicx}
\usepackage{textcomp}
\usepackage{hyperref}
\usepackage{tabularx}
\newcommand\BibTeX{{\rmfamily B\kern-.05em \textsc{i\kern-.025em b}\kern-.08em
T\kern-.1667em\lower.7ex\hbox{E}\kern-.125emX}}
\definecolor{orcidlogocol}{HTML}{A6CE39}
\tikzset{
  orcidlogo/.pic={
    \fill[orcidlogocol] svg{M256,128c0,70.7-57.3,128-128,128C57.3,256,0,198.7,0,128C0,57.3,57.3,0,128,0C198.7,0,256,57.3,256,128z};
    \fill[white] svg{M86.3,186.2H70.9V79.1h15.4v48.4V186.2z}
                 svg{M108.9,79.1h41.6c39.6,0,57,28.3,57,53.6c0,27.5-21.5,53.6-56.8,53.6h-41.8V79.1z M124.3,172.4h24.5c34.9,0,42.9-26.5,42.9-39.7c0-21.5-13.7-39.7-43.7-39.7h-23.7V172.4z}
                 svg{M88.7,56.8c0,5.5-4.5,10.1-10.1,10.1c-5.6,0-10.1-4.6-10.1-10.1c0-5.6,4.5-10.1,10.1-10.1C84.2,46.7,88.7,51.3,88.7,56.8z};
  }
}

\newcommand\orcidiconKLW[1]{\href{https://orcid.org/0000-0002-XXXX-YYYY}{\mbox{\scalerel*{
\begin{tikzpicture}[yscale=-1,transform shape]
\pic{orcidlogo};
\end{tikzpicture}
}{|}}}}

\newcommand\orcidiconFGS[1]{\href{https://orcid.org/0000-0002-5090-9007}{\mbox{\scalerel*{
\begin{tikzpicture}[yscale=-1,transform shape]
\pic{orcidlogo};
\end{tikzpicture}
}{|}}}}
\newcommand\orcidiconAAS[1]{\href{https://orcid.org/0000-0002-XXXX-YYYY}{\mbox{\scalerel*{
\begin{tikzpicture}[yscale=-1,transform shape]
\pic{orcidlogo};
\end{tikzpicture}
}{|}}}}

\title{\LARGE \bf
A Virtual-Variable-Length method for robust Inverse Kinematics of multi-segment continuum robots}

 \author{Weiting Feng$^{1 }$, Federico Renda$^{2}$, Yunjie Yang$^{1}$ and Francesco Giorgio-Serchi$^{1 \orcidiconFGS{0000-0002-5090-9007}}$
 \thanks{$^{1}$Weiting Feng, Yunjie Yang and Francesco Giorgio-Serchi are with the School of Engineering, University of Edinburgh, Edinburgh, U.K. (Correspondence: {\tt\small f.giorgio-serchi@ed.ac.uk}).} 
 \thanks{$^{2}$Federico Renda is with Department of Mechanical Engineering, Khalifa University, Abu Dhabi, UAE.} 
 }

\begin{document}

\maketitle
\thispagestyle{empty}
\pagestyle{empty}

\begin{abstract}
This paper proposes a new, robust method to solve the inverse kinematics (IK) of multi-segment continuum manipulators. 
Conventional Jacobian-based solvers, especially when initialized from neutral/rest configurations, often exhibit slow convergence and, in certain conditions, may fail to converge (deadlock). 
The Virtual-Variable-Length (VVL) method proposed here introduces fictitious variations of segments' length during the solution iteration, conferring virtual axial degrees of freedom that alleviate adverse behaviors and constraints, thus enabling or accelerating convergence. 
Comprehensive numerical experiments were conducted to compare the VVL method against benchmark Jacobian-based and Damped Least Square IK solvers. 
Across more than 1.8$\times$10$^6$ randomized trials covering manipulators with two to seven segments, the proposed approach achieved up to a 20\% increase in convergence success rate over the benchmark and a 40–80\% reduction in average iteration count under equivalent accuracy thresholds (10$^{-4}$–10$^{-8}$). 
While deadlocks are not restricted to workspace boundaries and may occur at arbitrary poses, our empirical study identifies boundary-proximal configurations as a frequent cause of failed convergence and the VVL method mitigates such occurrences over a statistical sample of test cases. 
\end{abstract}

\section{Introduction}
\label{sec: intro}

Soft robotics has emerged as a promising field that bridges the gap between traditional rigid robotics and biological systems. By leveraging compliant, deformable materials, soft robots can exhibit continuous deformations, adapt safely to unstructured environments, and achieve tasks that are difficult or impossible for rigid manipulators, such as delicate object handling and navigation in confined spaces \cite{Rus2015, Trivedi2008}. Their inherent compliance and continuum morphology also provide a natural robustness to uncertainties, making them particularly suitable for human–robot interaction, biomedical applications and operation in aquatic environments \cite{Kim2013, Walker2025Robosoft}.

Among the various modeling frameworks for soft continuum manipulators, the Constant Curvature (CC) model has become one of the most popular and tractable approaches \cite{Walker2013}. Despite its strong assumptions and limitations, it remains a staple in the modeling \cite{Grassmann2025} and control \cite{DellaSantina2023, WalkerRoboSoft} of soft manipulators. The CC assumption simplifies the complex continuous deformation of a soft segment into a finite set of curvature parameters, enabling efficient kinematic and dynamic modeling while maintaining reasonable accuracy for many practical applications \cite{Webster2010}. This reduction allows analytical expressions for forward and inverse kinematics (IK), and most importantly it allows a direct translation towards canonical control strategies originally designed for rigid-links manipulators, making it an attractive choice for real-time control and path planning in continuum and cable-driven soft manipulators.

In analogy with traditional manipulators, IK of multi-segment CC-based continuum manipulators can be executed via canonical Jacobian-based iterative methods. These methods exploit the differential relationship between the end-effector pose and the configuration parameters to iteratively update the manipulator state toward the desired target \cite{LynchBook}. Compared with global optimization or sampling-based techniques, Jacobian methods are computationally lightweight and well-suited for online control, especially when combined with damped least-squares or adaptive step-size strategies to handle near-singular configurations \cite{Buss2004}.

However, traditional Jacobian methods in CC models suffer from limitations which are not commonly discussed, but frequently encountered. One such issue is the sensitivity to singularities, where the Jacobian matrix becomes ill-conditioned or loses rank, leading to unstable or divergent iterations \cite{Buss2004, Selig2010}. Moreover, a less-discussed but practically significant problem arises in certain iterative processes: a phenomenon commonly referred to as the \textit{deadlock} effect \cite{Aristidou2016, Kolpashchikov2018}. This effect manifests itself both in rigid-link and CC models: in the latter case one or more segments of the manipulator may fold onto themselves, forming a circular configuration with a bending angle exceeding \(2 \pi\), which causes the manipulator to coil up and fail to recover its straightened posture. This phenomenon not only prevents convergence but can also lead to physically infeasible configurations that violate the CC assumption.

In addition, existing IK formulations for continuum manipulators are often characterized by a pronounced sensitivity to the choice of initial conditions. The convergence behavior of most numerical or iterative IK solvers is strongly influenced by the proximity of the initial guess to the true solution, with poor initialization frequently resulting in divergence or convergence to physically infeasible configurations. This sensitivity is further exacerbated in manipulators composed of multiple constant curvature segments, wherein the dimensionality and nonlinearity of the configuration space increase substantially, thereby reducing the likelihood of obtaining a convergent and accurate inverse solution.

In this work, we propose a novel Jacobian-based approach for CC models that effectively mitigates the aforementioned problems of \textit{deadlock} and sensitivity to the initial guess, while enhancing the overall convergence performance. The core idea of the proposed solution method is to introduce a \textit{Virtual Variable Length} (VVL) segment, which iteratively adjusts during the iteration process. By allowing the segment length to vary virtually, the algorithm gains an additional degree of freedom that helps the solver escape geometrical constraints, \textit{deadlock} coiled configurations and navigate around local singularities. This modification leads to both faster convergence and a higher success rate in achieving valid inverse kinematic solutions. Extensive numerical experiments demonstrate that our method significantly improves convergence robustness compared to traditional Jacobian schemes, especially in highly curved configurations or near-singular regions.


\section{Inverse Kinematics of Continuum Robots}
\label{sec: Methods}

\subsection{Kinematics of Constant Curvature manipulators}

\begin{figure}
    \centering
    \includegraphics[width=0.5\linewidth]{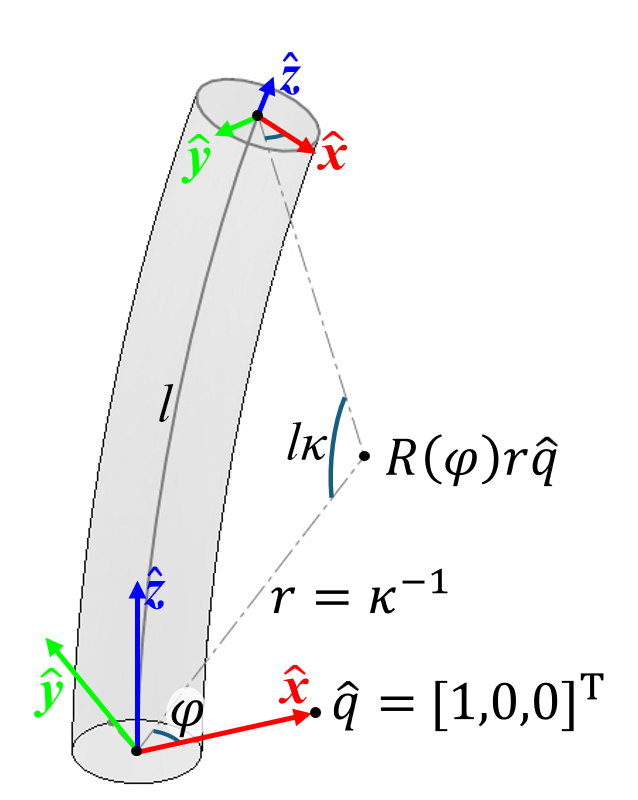}
    \caption{Schematic of Constant Curvature segment, as in \cite{Allen2020}.}
    \label{fig:schematic 01}
\end{figure}

Here, we concern ourselves with CC models of multi-segment manipulators and formulate a well-established Jacobian-based recursive algorithm for the solution of the IK problem of CC continuum manipulators. We formulate the kinematics of such a system using exponential coordinates and taking as reference the Fig.\ref{fig:schematic 01}  Following CC assumption and considering a segment of the manipulator of reference length $l$, the deformation of the manipulator within each segment is a pure bending. Therefore, we establish a coordinate frame at the base of this segment, and denote the bending direction by an angle $\varphi$ with respect to the $x$-axis, and the curvature by $\kappa$.  
The transformation from the base to the tip of this segment can then be regarded as a rotation about an axis whose direction is given by  
\(
\omega' = R(\varphi)\,\omega,
\) 
where $R(\varphi)$ is the in-plane rotation within the $xy$-plane, and $\omega = [0,\,1,\,0]^T$ is a unit vector along the $y$-axis, serving as the reference axis when $\varphi=0$.  
This axis also passes through a point $p$ in space, whose location is determined jointly by the curvature $\kappa$ and the bending direction $\varphi$. This form is a special case of the representations presented in \cite{Renda2018_TRO} and \cite{Anup2025}.
From geometric considerations, we obtain  
\(
p = R(\varphi)\, r \hat{q},
\) 
where $r=\kappa^{-1}$ is the distance from the segment center to the rotation axis, and $\hat{q}=[1,\,0,\,0]^T$ is a unit vector indicating the reference direction of $p$ when $\varphi=0$.  

The rotation about this axis can be represented by a twist,  
\begin{equation}
\mathcal{V}'
=
\left[\begin{matrix}
\omega'\\
-^{*}\!\omega'p
\end{matrix}\right] 
=\left[\begin{matrix}
 R(\varphi) \omega\\
-^{*}\!(R(\varphi) \omega) rR(\varphi)\hat{q}
\end{matrix}\right] 
=
\left[\begin{matrix}
 R(\varphi) \omega\\
 rR(\varphi)^{*}\!\hat{q} \omega
\end{matrix}\right] 
\end{equation}
where ${}^{*}(\cdot)$ stands for the skew symmetric form, as per \ref{sec: appendix}.  
It is worth noticing that this rotation differs from the case of a rigid link rotating about a fixed axis, since here the rotation angle $\theta = l\kappa$ and the axis offset vector $p$ are intrinsically coupled.  
Therefore, we directly define a finite-rotation twist that explicitly incorporates $\theta$, and by adding the subscript to denote each segment, we obtain:  
\begin{equation}
\mathcal{V}_i=l_i  \left[\begin{matrix}
\kappa_i R(\varphi_i) \omega\\
R(\varphi_i)  ^{*}\!\hat{q} \omega
\end{matrix}\right] 
\end{equation}

\noindent which expresses the fact that the strain twist $\mathcal{V}_i$ of each segment depends on the two variables $\kappa_i$ and $\varphi_i$, \cite{Allen2020}. We can multiply the exponential of each 
strain twist together and get the forward kinematic equation:
\begin{equation}\label{FK}
T_e=e^{^{*}\!\mathcal{V}_1}e^{^{*}\!\mathcal{V}_2}...e^{^{*}\!\mathcal{V}_n}
\end{equation}

\subsection{IK via benchmark Jacobian method}
\label{subsec: Inverse Kinematics of CC Model with Screw Theory}

In order to proceed with calculating the IK, we borrow from the traditional approach commonly employed for rigid links, where we first calculate the Jacobian matrix by partial differentiation of the forward kinematic function (\ref{FK}).
\begin{equation}\label{dT}
\renewcommand{\arraystretch}{1.2}
\setlength{\arraycolsep}{1.5pt}
\begin{array}{r l}
dT_e=&\text{ }\text{ }\text{ }(\partial_{\kappa_1}e^{^{*}\!\mathcal{V}_1} d\kappa_1+\partial_{\varphi_1}e^{^{*}\!\mathcal{V}_1} d\varphi_1 )
e^{^{*}\!\mathcal{V}_2}...e^{^{*}\!\mathcal{V}_n}\\
&+
e^{^{*}\!\mathcal{V}_1}(\partial_{\kappa_2}e^{^{*}\!\mathcal{V}_2} d\kappa_2+\partial_{\varphi_2}e^{^{*}\!\mathcal{V}_2}d\varphi_2 )...e^{^{*}\!\mathcal{V}_n}\\
&+
e^{^{*}\!\mathcal{V}_1}e^{^{*}\!\mathcal{V}_2}...e^{^{*}\!\mathcal{V}_{n-1}}(\partial_{\kappa_n}e^{^{*}\!\mathcal{V}_n} d\kappa_n+\partial_{\varphi_n}e^{^{*}\!\mathcal{V}_n} d\varphi_n )
\end{array}
\end{equation}
However, unlike the case encountered for rigid robots, because $\partial_{\kappa_i}\!\!\!{^{*}\!\mathcal{V}_i}$, $\partial_{\varphi_i}\!\!\!{^{*}\!\mathcal{V}_i}$ and $\mathcal{V}_i$ usually do not commute, then $(\partial_{\kappa_1}\!\!\!{^{*}\!\mathcal{V}_1})e^{^{*}\!\mathcal{V}_1}$ is usually not equal to $\partial_{\kappa_1}e^{^{*}\!\mathcal{V}_1}$. To address this issue, we define the following (as in \cite{Renda2017JMR}):
\begin{equation}\label{not_commuting_terms}
^{*}\!\mathcal{V}^\kappa_ie^{^{*}\!\mathcal{V}_i}=\partial_{\kappa_i}e^{^{*}\!\mathcal{V}_i},\quad 
^{*}\!\mathcal{V}^\varphi_ie^{^{*}\!\mathcal{V}_i}=\partial_{\varphi_i}e^{^{*}\!\mathcal{V}_i}
\end{equation}
The expansion of $^{*}\!\mathcal{V}^\kappa_i$ and $^{*}\!\mathcal{V}^\varphi_i$, see \ref{sec: appendix}.
By substituting these terms in (\ref{not_commuting_terms}) into (\ref{dT}) and then dividing by (\ref{FK}) we get:
\begin{equation}
\renewcommand{\arraystretch}{1.2}
\setlength{\arraycolsep}{1.5pt}
\begin{array}{r l}
dT_eT_e^{-1}=&(^{*}\!\mathcal{V}^\kappa_1 d\kappa_1+{}^{*}\!\mathcal{V}^\varphi_1 d\varphi_1 )\\
&+\text{Ad}_{e^{^{*}\!\mathcal{V}_1}}(^{*}\!\mathcal{V}^\kappa_2 d\kappa_2+{}^{*}\!\mathcal{V}^\varphi_2 d\varphi_2 )\\
&+\text{Ad}_{e^{^{*}\!\mathcal{V}_1}e^{^{*}\!\mathcal{V}_2}\cdots e^{^{*}\!\mathcal{V}_{n-1}}}(^{*}\!\mathcal{V}^\kappa_n d\kappa_n+{}^{*}\!\mathcal{V}^\varphi_n d\varphi_n )
\end{array}
\end{equation}
Where $\text{Ad}_T(\cdot)=T(\cdot)T^{-1}$ is the adjoint action of the lie group element $T$.

Let $^{*}\!\mathcal{V}_e = \text{ln}(T_e)$, and transform the format of this equation from twist matrix to twist vector:
\begin{equation}
\renewcommand{\arraystretch}{1.2}
\setlength{\arraycolsep}{1.5pt}
\begin{array}{r l}
d\mathcal{V}_e=&(\mathcal{V}^\kappa_1 d\kappa_1+\mathcal{V}^\varphi_1 d\varphi_1 )+\boldsymbol{Ad}_{e^{^{*}\!\mathcal{V}_1}}(\mathcal{V}^\kappa_2 d\kappa_2+\mathcal{V}^\varphi_2 d\varphi_2 )\\
&+\boldsymbol{Ad}_{e^{^{*}\!\mathcal{V}_1}e^{^{*}\!\mathcal{V}_2}...e^{^{*}\!\mathcal{V}_{n-1}}}(\mathcal{V}^\kappa_n d\kappa_n+\mathcal{V}^\varphi_n d\varphi_n )  
\end{array}
\end{equation}
where $\boldsymbol{Ad}_T$ is the adjoint representation of $T$, defined as per \ref{sec: appendix}. 
Therefore the Jacobian matrix can be written as:

\begin{equation}\label{jacobian}
\begin{aligned}
J &= [\begin{matrix}
\mathcal{V}^\kappa_1 & \mathcal{V}^\varphi_1 &
\boldsymbol{Ad}_{e^{^{*}\!\mathcal{V}_1}}\!\!\mathcal{V}^\kappa_2 &
\boldsymbol{Ad}_{e^{^{*}\!\mathcal{V}_1}}\!\!\mathcal{V}^\varphi_2 & \cdots
\end{matrix} \\[4pt]
&\quad \begin{matrix}
\boldsymbol{Ad}_{e^{^{*}\!\mathcal{V}_1}e^{^{*}\!\mathcal{V}_2}\cdots e^{^{*}\!\mathcal{V}_{n-1}}}\!\!\mathcal{V}^\kappa_n &
\boldsymbol{Ad}_{e^{^{*}\!\mathcal{V}_1}e^{^{*}\!\mathcal{V}_2}\cdots e^{^{*}\!\mathcal{V}_{n-1}}}\!\!\mathcal{V}^\varphi_n
\end{matrix}]
\end{aligned}
\end{equation}

This form of the Jacobian matrix is a employed recursive formula to solve a converged IK solution numerically as follows,
\begin{equation}\label{jacobian-method}
\begin{aligned}
\Delta x &=
\left[
\begin{matrix}
    \Delta\kappa_1 & \Delta\varphi_1 & \cdots &
    \Delta\kappa_n & \Delta\varphi_n
\end{matrix}
\right]^\top \\
&= {\beta} J^+ \mathcal{V}_{D}
\end{aligned}
\end{equation}

\noindent where $T_D$ is the target posture, $(\cdot)^+$ is the pseudo-inverse of the matrix, $\beta$ is an iteration step size control factor, and \(^{*}\!\mathcal{V}_{D} = \ln(T_D T_e^{-1})\) is the twist from the current posture of end-effector to the desired posture in current frame. The recursive solution technique shown in (\ref{jacobian-method}) is formally no different from the traditional approaches encountered in rigid-link systems, \cite{LynchBook}.

\subsection{IK via damped least-squares method}
\label{subsec: The DLS Method}
As a comparison with the proposed VVL approach and the benchmark Jacobian method, the damped least-squares (DLS) method in \cite{doi:10.1137/0715063} is also considered.
The update is computed as
\begin{equation}\label{DLS method}
\begin{aligned}
\Delta x &=
\left[
\begin{matrix}
    \Delta\kappa_1 & \Delta\varphi_1 & \cdots &
    \Delta\kappa_n & \Delta\varphi_n
\end{matrix}
\right]^\top \\
&= {\beta} J^\top \left( J J^\top + \lambda^2 I \right)^{-1} \mathcal{V}_{D}
\end{aligned}
\end{equation}
where $\lambda>0$ is the damping factor.
The regularization term $\lambda^2 I$ improves numerical conditioning near singular configurations and prevents excessively large joint updates.

\subsection{IK via Virtual-Variable-Length method}
\label{subsec: The Augment Jacobian Method}

The core concept underlying the proposed VVL IK solution method is the introduction of a fictitious variable segment length, which is adaptively adjusted throughout the iterative solving process. In conventional formulations, the physical segment lengths of a piecewise constant curvature manipulator are fixed, thereby constraining the solver to a limited configuration space during the optimization process. By contrast, in the proposed approach, each segment is temporarily permitted to vary its effective length as a virtual parameter during the iteration. This relaxation introduces an additional degree of freedom that enables the solver to circumvent restrictive geometric constraints and other unwanted byproducts of the recursive calculation. Once convergence toward a feasible configuration is achieved, the virtual segment lengths are smoothly restored towards their nominal physical values, yielding a consistent and physically valid IK solution.

We provide a formal description of the VVL algorithm by first incorporating the virtual variable lengths element into the iterative solution formula of (\ref{jacobian}).  
Following an analogous treatment as previously shown for $\kappa_i$ and $\varphi_i$ and expanding based on an analogous derivation in \cite{Renda2017JMR}, we finally obtain,
\begin{equation}
\begin{aligned}
J' &= [\begin{matrix}
\mathcal{V}^\kappa_1&\mathcal{V}^\varphi_1&\mathcal{V}^l_1&\cdots\boldsymbol{Ad}_{e^{^{*}\!\mathcal{V}_1}e^{^{*}\!\mathcal{V}_2}...e^{^{*}\!\mathcal{V}_{n-1}}}\!\!\mathcal{V}^\kappa_n 
\end{matrix} \\[4pt]
&\quad \begin{matrix}
&\boldsymbol{Ad}_{e^{^{*}\!\mathcal{V}_1}e^{^{*}\!\mathcal{V}_2}...e^{^{*}\!\mathcal{V}_{n-1}}}\!\!\mathcal{V}^\varphi_n    
&\boldsymbol{Ad}_{e^{^{*}\!\mathcal{V}_1}e^{^{*}\!\mathcal{V}_2}...e^{^{*}\!\mathcal{V}_{n-1}}}\!\!\mathcal{V}^l_n 
\end{matrix}]
\end{aligned}
\end{equation}

\noindent which allows us to reformulate the Jacobian inclusive of the added variable-length parameter. This can be implemented in a modified version of (\ref{jacobian-method}) to obtain the iterative step of the VVL method:
\begin{equation}\label{VVL_method}
\begin{aligned}
\Delta x' &=
\left[
\begin{matrix}
    \Delta\kappa_1 & \Delta\varphi_1 & \Delta l_1 & \cdots &
    \Delta\kappa_n & \Delta\varphi_n & \Delta l_n
\end{matrix}
\right]^\top \\
&= \beta (J')^+ \mathcal{V}_{D}
\end{aligned}
\end{equation}
Successful execution of the calculation and phsyical consistency dictates that, after the iteration converges to the target pose $T_D$, the segments' length $l_i$ return to their original physical values, denoted as $L_i$. To achieve this, we impose the following constraint during the recursive calculation process: 
\(
\Delta l_i = L_i - l_i
\) 
which inherently drives the final values of $l_i$ to the expected one.  
Notice that if this correction stage were applied as a secondary iteration on top of the original scheme, it would alter the original convergence behavior.

To avoid this, we instead incorporate the correction directly into the primary iteration process by making use of an \textit{augmented} Jacobian $J''$, yielding:  
\begin{equation}
\Delta x'=(J'')^+\left[\begin{matrix}
\mathcal{V}_{D}\\
L-l
\end{matrix}\right]
\end{equation}
where
\begin{equation}
J''=\left[\begin{matrix}
J'\\
P
\end{matrix}\right], \quad
L = \left[\begin{smallmatrix}L_1\\L_2\\\vdots\\L_n\end{smallmatrix}\right]
, \quad
l = \left[\begin{smallmatrix}l_1\\l_2\\\vdots\\l_n\end{smallmatrix}\right]
\end{equation}
Here $P$ is a matrix composed of $0$s and $1$s, which corresponds to the elements of the vectors on both sides of the equation that represent the length of the segments.

For the sake of compactness, this can be rearranged by adjusting the order of the elements in $x'$ and $J'$ to obtain the following.
\begin{equation}
x_\text{new}=\left[\begin{matrix}
\kappa\\\varphi\\l
\end{matrix}\right],\quad
\kappa=\left[\begin{smallmatrix}
\kappa_1\\\kappa_2\\\vdots\\\kappa_n
\end{smallmatrix}\right],\quad
\varphi=\left[\begin{smallmatrix}
\varphi_1\\\varphi_2\\\vdots\\\varphi_n
\end{smallmatrix}\right]
\end{equation}
and
\begin{equation}
\begin{aligned}
\renewcommand{\arraystretch}{1.2}
\setlength{\arraycolsep}{3pt}
J_\text{new}=&\left[\begin{matrix}
J_\kappa&J_\varphi&J_l
\end{matrix} 
\right]
\end{aligned}
\end{equation}
Where
\begin{equation}\
\begin{matrix}
J_\alpha=\left[\begin{matrix}
\mathcal{V}^\alpha_1&\cdots&\boldsymbol{Ad}_{e^{^{*}\!\mathcal{V}_1}e^{^{*}\!\mathcal{V}_2}...e^{^{*}\!\mathcal{V}_{n-1}}}\!\!\mathcal{V}^\alpha_n 
\end{matrix} 
\right]\\
\text{and }\alpha = \kappa,\varphi,l
\end{matrix}
\end{equation}
In this case:
\begin{equation}
\renewcommand{\arraystretch}{1.2}
\setlength{\arraycolsep}{3pt}
P_\text{new} = \left[\begin{matrix}\mathbf{0}&I\end{matrix} 
\right], \text{and } J_\text{a} = \left[\begin{matrix}
J_\text{new}\\
P
\end{matrix}\right]
\end{equation}

\begin{equation}
\Delta x_\text{new}=\beta (J_\text{a})^+\left[\begin{matrix}
\mathcal{V}_{D}\\
L-l
\end{matrix}\right]
\end{equation}

\begin{figure*}[ht!!!]
    \centering
   \includegraphics[width=0.95\linewidth]{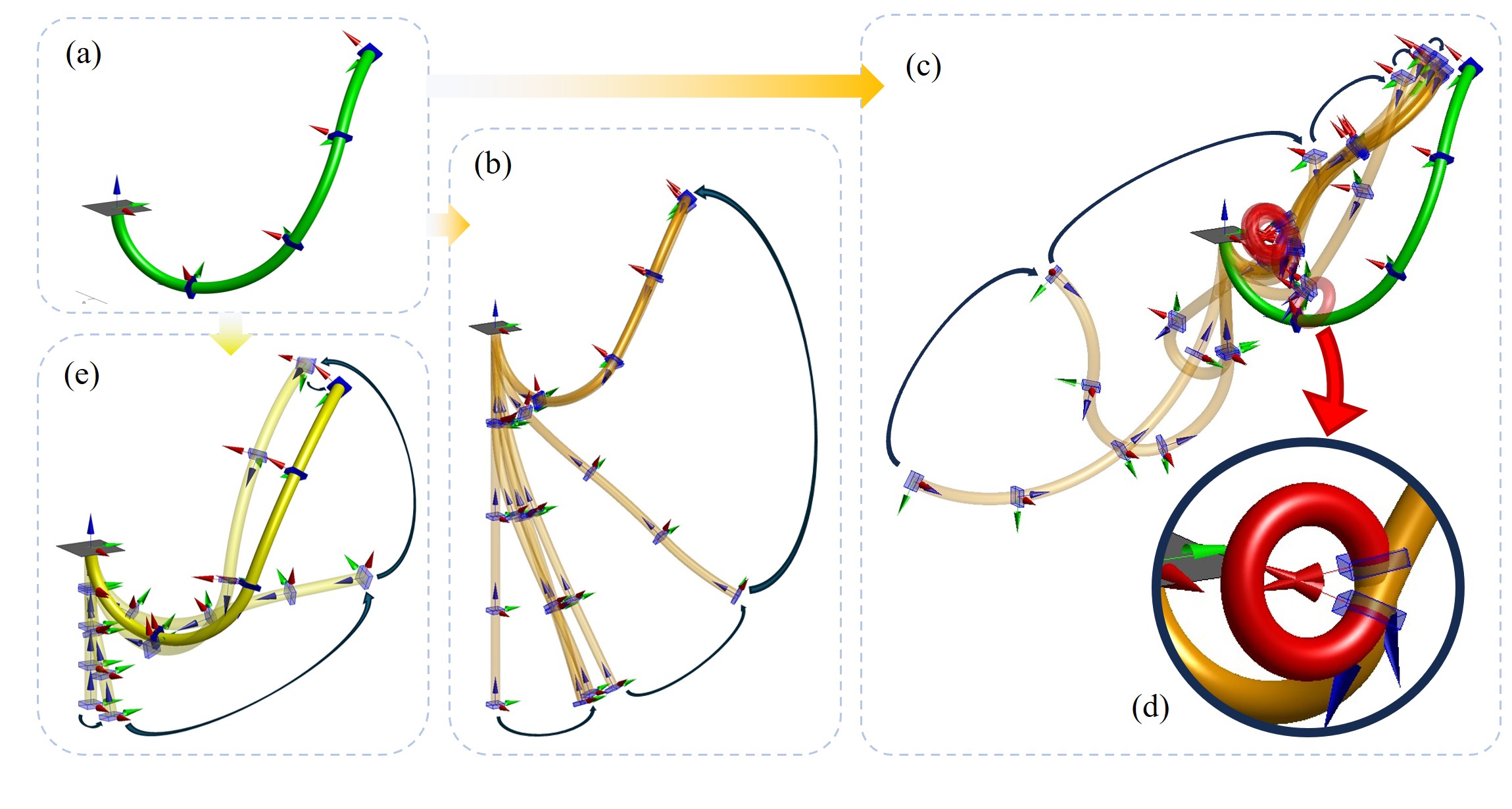}
    \caption{
    Illustration of convergence progression to desired configuration (a) identified with a green color, from two distinct initial guesses (b) and (c) when using a traditional Jacobian method. In (b) a solution is obtained after 81 iterations, as demonstrated by the final configuration overlapping with the reference state. In (c) a solution is not reached due to a \textit{deadlock} configuration, shown in red in (d). In (e), solution via the proposed VVL method is reached within 30 iterations; notice the initial variable segments' length which aid in the convergence process.}
    \label{fig:mechanism}
\end{figure*}

Finally, the VVL method is initiated with a perturbed initial guess where the starting length of the segments are virtually shorter than the original ones.

To assess numerical accuracy of the IK algorithms we make use of the magnitude of the expression \(\mathcal{V}_{D}\), which reflects the mismatch between the current and the target configurations.  
Therefore, \(e_\mathcal{V} = |\mathcal{V}_{D}|\) can be regarded as a composite error that simultaneously accounts for both orientation and position discrepancies.  
Its physical meaning can be interpreted from the following expression:
\begin{equation}
    \begin{aligned}
    e_\mathcal{V}^2 &= 
    \dot\theta_e^2 \left[\begin{matrix}
        \omega_e^T  & (\omega_e \times q_e + h \omega_e)^T
    \end{matrix}\right]
       \left[\begin{matrix}
       \omega_e  \\ \omega_e \times q_e + h \omega_e
    \end{matrix}\right]\\
    &= \dot\theta_e^2 + \dot\theta_e^2 h^2 + \dot\theta_e^2 (\omega_e \times q_e)^2\\
    &\approx \Delta\theta_e^2 + \Delta x_\parallel^2 + \Delta x_\perp^2
    \end{aligned}
\end{equation}
where \(\omega_e\), \(\dot{\theta}_e\), \(q_e\) and \(h\) define the geometrical parameterization of a screw as per \cite{LynchBook}. Consequently, the cumulative error depends on the angular error \(\Delta\theta_e^2\), the axial translational error \(\Delta x_\parallel^2\), and the circumferential (perpendicular) translational error \(\Delta x_\perp^2\) of the screw. The approximation
\begin{equation}\label{tolerance}
e_\mathcal{V} \approx \sqrt{\Delta\theta_e^2 + \Delta x_\parallel^2 + \Delta x_\perp^2}
\end{equation}
provides an integrated measure of pose discrepancy in both rotation and translation which we employ in the remainder of the text.

\begin{figure*}[t]
    \centering
    \includegraphics[trim= 5 10 5 5,clip, width=0.8\linewidth]{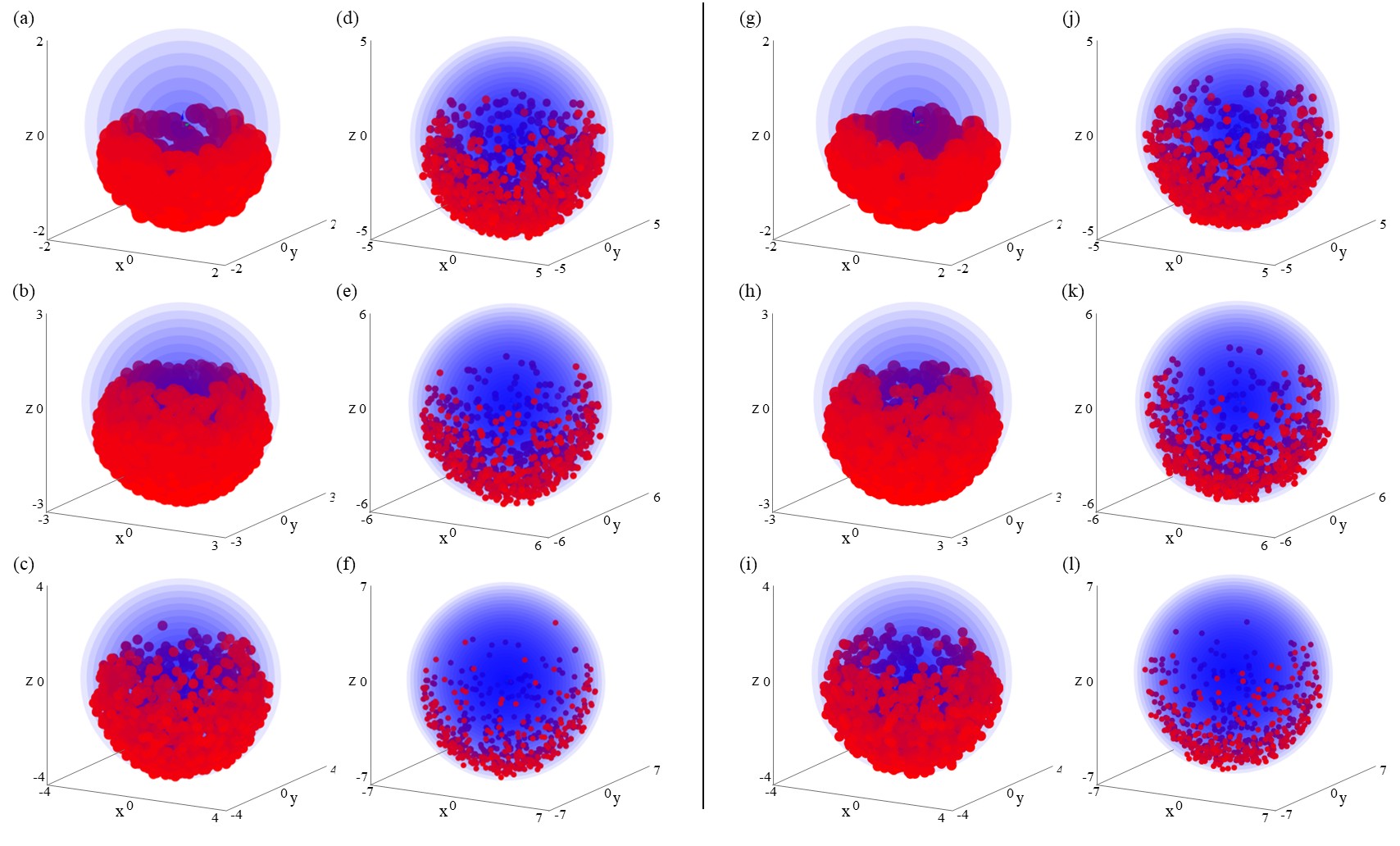}
    \caption{Workspace of the continuum manipulator (blue) and failed IK solutions (red) obtained with Jacobian method (a)-(f) and DLS method (g)-(l).
    Subfigures (a) and (g) correspond to the two-segment manipulator,
    (b) and (h) to a three-segment manipulator,
    (c) and (i) to a four-segment manipulator,
    (d) and (j) to a five-segment manipulator,
    (e) and (k) to a six-segment manipulator,
    and (f) and (l) to a seven-segment manipulator.
    Subfigures (a)–(f) are obtained using the benchmark Jacobian method described in Sec.~\ref{subsec: Inverse Kinematics of CC Model with Screw Theory},
    while (g)–(l) are obtained using the DLS method described in Sec.~\ref{subsec: The DLS Method}.}
    \label{fig:O}
\end{figure*}

\begin{figure*}
    \centering
    \includegraphics[width=0.8\linewidth]{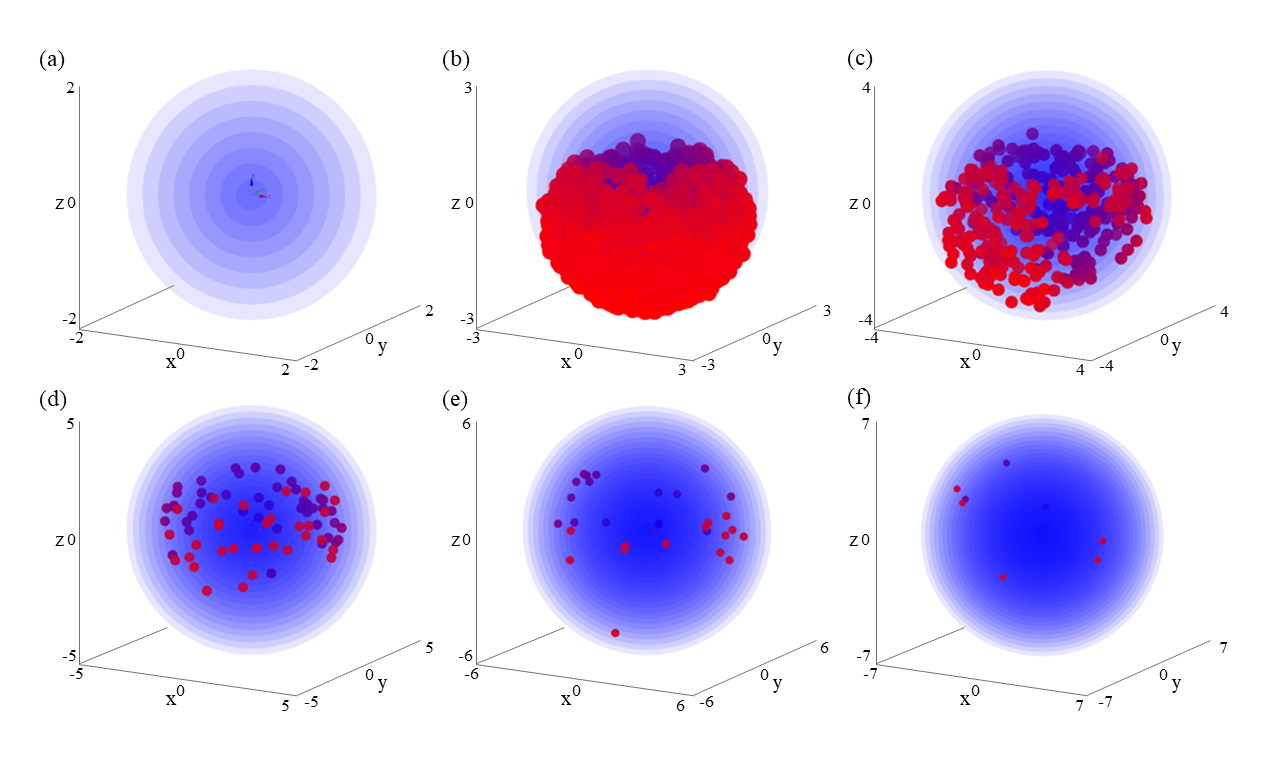}
    \caption{Workspace of the continuum manipulator, in blue, and failed IK solutions in red obtained with the VVL method (sec.\ref{subsec: The Augment Jacobian Method}), respectively for the case of a manipulator made of (a) two segments, (b) three segments, (c) four segments, (d) five segments, (e) six segments and (f) seven segments.}
    \label{fig:V}
\end{figure*}

\section{Results}
\label{sec: Results}

\subsection{Validation of the VVL method}
\label{subsec: validation}
A demonstration of how the VVL algorithm works and how it compares to the traditional Jacobian method is shown in Fig. \ref{fig:mechanism}.
Here subfigure (a) shows the desired configuration of a continuum robot composed of four segments, each of equal length \( l = 1 \).  
Its current configuration can be expressed by the forward kinematics, where the curvature and orientation angles of each segment are given as
\[
\boldsymbol{\kappa} = 
\begin{bmatrix}
\pi/2 \\[3pt]
\pi/3 \\[3pt]
\pi/4 \\[3pt]
\pi/5
\end{bmatrix},
\qquad
\boldsymbol{\phi} =
\begin{bmatrix}
\pi/5 \\[3pt]
\pi/4 \\[3pt]
\pi/2 \\[3pt]
3\pi/4
\end{bmatrix}.
\]
Subfigures (b) and (c) illustrate the iterative processes of IK using the traditional Jacobian method when starting from two different initial guesses: in (b) the manipulator is initialized from a natural rest configuration (as if subject to gravity). Each configuration in insets (b) and (c) corresponds to a snapshot taken every 6 iterative steps. It can be observed that in case (b), the solution converges to the desired configuration with an error smaller than \(10^{-8}\) after 81 iterations. Approximately 30 iterations are required to escape the initial singular region, and another 30 iterations are performed near the target configuration because the target lies close to the workspace boundary. In case (c), the algorithm fails to achieve further convergence after 60 iterations, with a remaining residual error of \(0.304\). This failure is due to the solution falling into a \textit{deadlock}, marked in red and highlighted in inset (d), that occurs at the 10th iteration step, resulting in a permanent loss of one segment’s effective length. The deadlock arises because the initial guess is oriented in the opposite direction to the target pose. During the iteration, each segment must bend appropriately so that the end-effector follows a feasible path toward the target. When the initial and target directions are opposite, the algorithm may attempt to bend a segment beyond a straight angle in order to move the end-effector from one side of that segment to the other. Such excessive bending makes the solution prone to producing a deadlock, preventing further convergence.


In Fig.~\ref{fig:mechanism}(e), the iterative process following the VVL method is illustrated. The initial guess configuration is perturbed with a starting segment length approximately one third of the actual length, as explained in section \ref{subsec: The Augment Jacobian Method}. After 30 iterations the solution successfully converges within the specified tolerance. Before introducing variable segment lengths, the manipulator at its gravitationally relaxed configuration exhibits a singularity, since all attainable motion directions are confined to the horizontal plane. By allowing the segment lengths to vary through the VVL formulation, an additional axial degree of freedom is introduced, effectively removing this singularity. The same principle also applies to singularities located near the boundary of the workspace. At the beginning of the process, each segment is stretched to reach the target configuration, causing the bending angles to become small during the early iterations and thus reducing the probability of encountering a deadlock. Although the conventional Jacobian method can in principle avoid singularities by starting from a randomly perturbed configuration instead of the neutral one, such initialization may still lead to deadlocks, as discussed above. In contrast, the proposed VVL approach enables the use of a neutral initial configuration, unaffected by singularities, thereby further decreasing the likelihood of deadlock.


\begin{figure}[t!!!]
    \centering
    \includegraphics[trim= 16 2 20 5,clip, width=0.85\columnwidth]{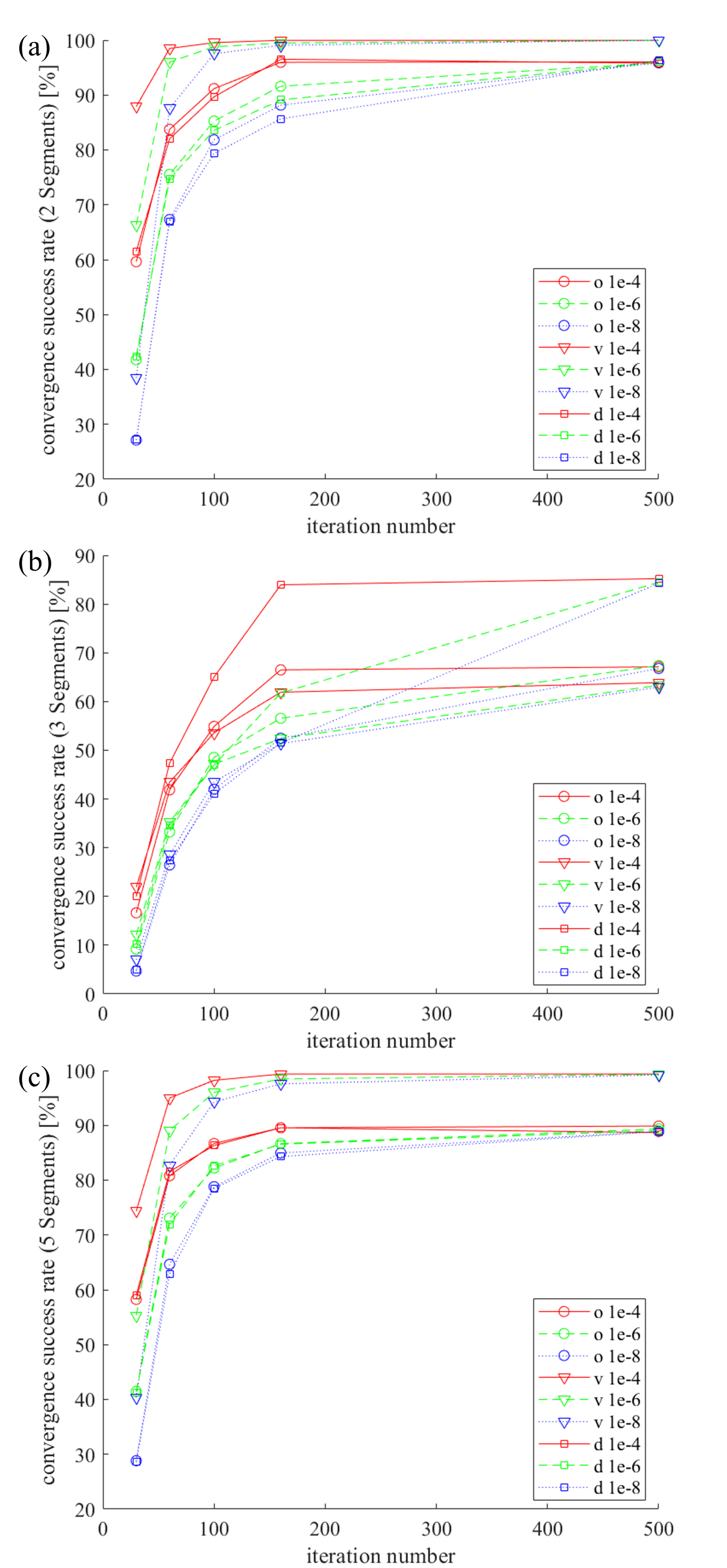}
    \caption{Convergence success rate with tolerances of 10$^{-4}$, 10$^{-6}$, 10$^{-8}$ for manipulators with 2, in (a), 3, in (b), and 5, in (c), segments respectively for the Jacobian method (indicated with the $o$ symbol in the legend), the VVL method (indicated by the $\text{v}$ symbol), and the DLS method (indicated by the $\text{d}$ symbol).}
    \label{fig:fig1}
\end{figure}

\subsection{VVL Computational Performance}

A series of numerical experiments were conducted to evaluate the performance of the proposed method in comparison with the conventional Jacobian-based approach and the damped least squares (DLS) method (\(\lambda = 0.01\)).
For each manipulator configuration ranging from two to seven segments, $15$ independent experimental groups were performed, each consisting of $10{,}000$ randomly sampled initial poses.
The experiments covered different iteration limits and convergence tolerances, as summarized by the parameter set
\(
(N_{\mathrm{iter}}, \varepsilon) \in \{30,\, 60,\, 100,\, 160,\, 500\} \times \{10^{-4},\, 10^{-6},\, 10^{-8}\}.
\)
The conventional Jacobian method, the proposed VVL method, and the DLS method were tested under identical conditions, yielding a total of
\(
3 \times 6 \times 5 \times 3 \times 10{,}000
\)
trials.
For each case, the proportion of trials that successfully converged to the target configuration within the specified limits was recorded and compared.

For the test in which the solution was allowed to converge over $500$ iterations with a convergence threshold of $10^{-4}$, we recorded the positions of all target poses.
These data are visualized in Fig.~\ref{fig:O} for the benchmark Jacobian method and the DLS method, Fig.~\ref{fig:V} for the VVL method. 
In these figures, all failed trials among the $10{,}000$ randomized runs for each configuration are marked as red points.
The failures are primarily attributed to the occurrence of \emph{deadlocks}, in which one segment effectively loses its active bending capability, preventing the manipulator from reaching certain target poses.
Consistent with our theoretical analysis, the DLS method does not significantly reduce the number of such failures for most segment numbers, indicating that deadlocks are structurally different from classical kinematic singularities and cannot be eliminated through damping regularization.
By contrast, the proposed VVL method substantially reduces the statistical occurrence of deadlocks, yielding a higher convergence success rate.

We also evaluated the accuracy of the proposed IK algorithms according to the tolerance metric of (~\ref{tolerance}).
The results are summarized in Fig.~\ref{fig:fig1}, where only the cases with two, three, and five segments are presented.
Each figure contains nine curves: three corresponding to the conventional Jacobian method, three to the DLS method, and three to the proposed VVL approach.
Within each group, the three curves represent convergence criteria of $10^{-4}$, $10^{-6}$, and $10^{-8}$.
With the exception of the three-segment case, the DLS method exhibits behavior similar to the conventional Jacobian approach, while the proposed VVL method persistently achieves higher success rates at lower iteration numbers.

\begin{figure}
    \centering
    \includegraphics[width=0.9\columnwidth]{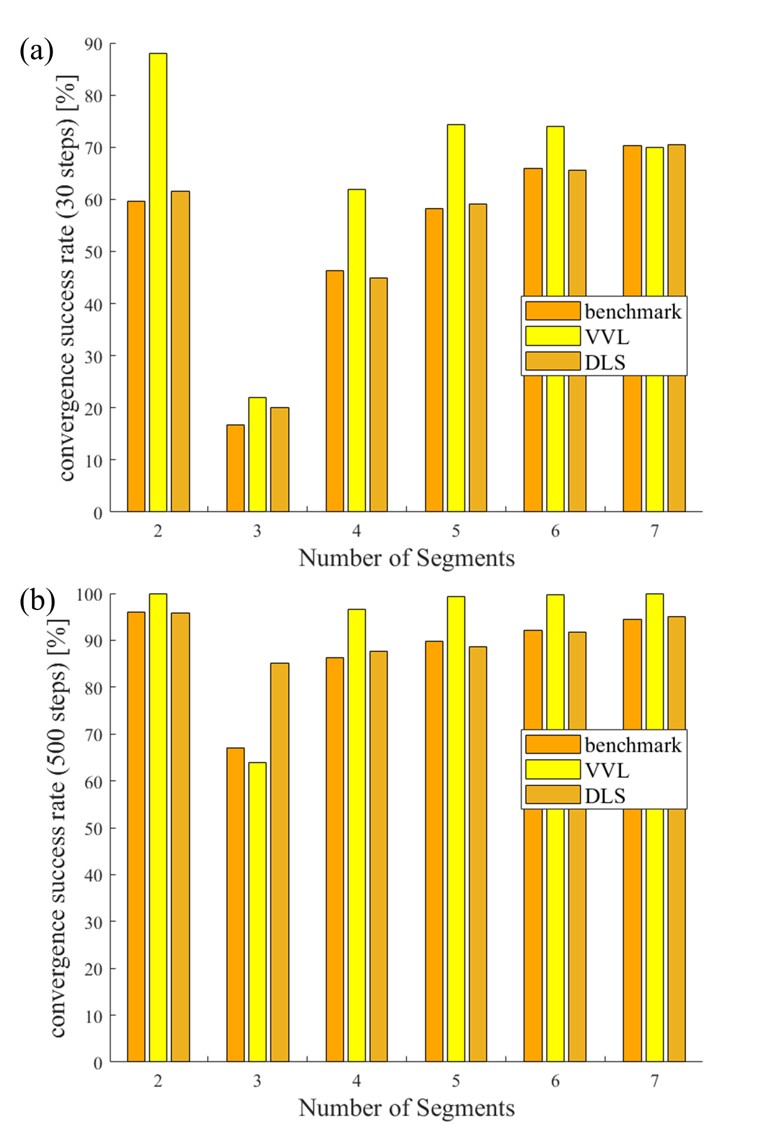}
    \caption{Convergence success rate for the benchmark Jacobian method (orange bar), the VVL method (yellow bar) and the DLS (brown bar) method with a tolerance of 10$^{-4}$ within (a) 30 iterations and (b) up to 500 iterations.}
    \label{fig:Fig}
\end{figure}

A more direct comparison is provided in Fig.~\ref{fig:Fig}, which summarizes the convergence success rates after $30$ and $500$ iterations across all segment numbers.
Once again, the three-segment case stands out as an outlier.
The peculiarity of the three-segment case
 can be explained by considering the ratio between the inner and outer workspace volumes: for three segments, the ratio is approximately \(4{:}5\). 
Since the target poses are uniformly sampled in the joint-angle space, the probability of selecting configurations within the inner region becomes higher than \(4/9\), thereby increasing the likelihood of early convergence stagnation. 
For manipulators with more segments, this ratio gradually increases (e.g., \(9{:}7\), \(16{:}9\), \(25{:}11\)), reducing this effect and resulting in a monotonic improvement in success rate with segment number.

Interestingly, in the three-segment case, the DLS method significantly improves the convergence success rate compared to the benchmark Jacobian approach and the VVL method.
Since the proposed VVL method is designed to eliminate boundary singularities, this result suggests that the convergence difficulty in the three-segment case cannot be attributed solely to workspace-surface singularities.
Instead, additional singular configurations must exist within the interior of the workspace, which are not avoidable through VVL-based strategies.

When the manipulator consists of only two segments, deadlocks rarely occur, and all three methods achieve relatively high success rates.
Nevertheless, the proposed VVL approach consistently attains higher convergence rates with fewer iterations.
Except for the three-segment case, where the improvement is comparatively moderate, the proposed method demonstrates a substantial enhancement in convergence success rate for all other configurations.




\section{Conclusions}
\label{sec: conclusions}

This work presents a virtual variable length (VVL) approach for solving the inverse kinematics of multi-segment continuum manipulators. 
By introducing virtual length variations during the iterative process, the proposed method mitigates boundary-related kinematic singularities while effectively overcoming structural deadlock in the inverse kinematics process.
Comprehensive numerical experiments demonstrate that the VVL approach consistently achieves higher convergence rates and significantly improves the overall robustness compared with the conventional Jacobian-based method and DLS method.

Nevertheless, the present study does not analytically characterize the precise mathematical mechanism underlying the occurrence of deadlocks. 
The explanation provided is based on extensive numerical observations and a probabilistic interpretation rather than a formal proof. 
Accordingly, the proposed method should be understood as a practical strategy that improves the probability of successful convergence rather than a guarantee of convergence in all cases. 
In rare instances, deadlocks may still occur.

The current implementation is based on the constant-curvature (CC) model and the underlying mathematical framework is formulated using screw theory. 
If the definition of the generalized velocity set \(\mathcal{V}'\) is extended beyond pure bending to include variable curvature components, the same principle can naturally be applied to models that include additional modes of strain and diverse variable-curvature formulations. 
Therefore, the proposed method offers a unified solution framework that can, in principle, be generalized to a broader class of continuum and soft robotic systems.

\section{Appendix}
\label{sec: appendix}
The skew symmetric matrix operator for 3-dimension vectors and 6-dimension twists:
\begin{equation}
v=\left[
\begin{smallmatrix}
    v_1\\
    v_2\\
    v_3
\end{smallmatrix}
\right] \rightarrow 
{}^*\!v=\left[
\begin{smallmatrix}
\rule{0.70em}{0pt}0 & -v_3 & \rule{0.70em}{0pt}v_2\\
\rule{0.70em}{0pt}v_3 & \rule{0.70em}{0pt}0 & -v_1\\
-v_2 & \rule{0.70em}{0pt}v_1 & \rule{0.70em}{0pt}0
\end{smallmatrix}
\right]
\end{equation}
\begin{equation}
\mathcal{V}=\left[
\begin{smallmatrix}
    \omega\\
    v
\end{smallmatrix}
\right] \rightarrow 
{}^*\!\mathcal{V}=\left[
\begin{smallmatrix}
{}^*\!\omega&\quad v\\
\mathbf{0}&\quad 0
\end{smallmatrix}
\right]
\end{equation}
Adjoint represent matrix of twists and homogeneous transformation matrices:
\begin{equation}
\mathcal{V}=
\left[\begin{smallmatrix}
\omega\\ v
\end{smallmatrix}\right] \rightarrow
\boldsymbol{ad}_{\mathcal{V}} =
\left[\begin{smallmatrix}
{}^*{\omega} & \boldsymbol{0}\\
 \quad {}^*v &\quad {}^*{\omega}
\end{smallmatrix}\right].
\end{equation}
\begin{equation}
T=
\left[\begin{smallmatrix}
R &\quad p\\ \mathbf{0} &\quad 1
\end{smallmatrix}\right] \rightarrow
\boldsymbol{Ad}_{T} =
\left[\begin{smallmatrix}
R & \quad \mathbf{0}\\
{}^{*}pR &\quad R
\end{smallmatrix}\right].
\end{equation}
Following from (\ref{not_commuting_terms}), we defined,
\begin{equation}
\renewcommand{\arraystretch}{1.2}
\setlength{\arraycolsep}{1.5pt}
\begin{array}{r l}
^{*}\!\mathcal{V}^\kappa_i&=\partial_{\kappa_i}e^{^{*}\!\mathcal{V}_i}e^{-^{*}\!\mathcal{V}_i}\\
&=(\partial_{\kappa_i}\sum_{n=0}^{\infty}\frac{1}{n!}{}^{*}\!\mathcal{V}_i^n)e^{-^{*}\!\mathcal{V}_i}
\end{array}
\end{equation}
Because $\partial_{\kappa_i}\!\!\!{^{*}\!\mathcal{V}_i^n}$ and ${}^{*}\!\mathcal{V}_i^n$ do not commute, instead of calculating it directly, we can use Hausdorff's formula and get:
\begin{equation}
^{*}\!\mathcal{V}^\kappa_i=\frac{1}{1!}\partial_{\kappa_i}\!\!\!{^{*}\!\mathcal{V}_i}+\frac{1}{2!}[\mathcal{V}_i,\partial_{\kappa_i}\!\!\!{^{*}\!\mathcal{V}_i}]+\frac{1}{3!}[\mathcal{V}_i,[\mathcal{V}_i,\partial_{\kappa_i}\!\!\!{^{*}\!\mathcal{V}_i}]]+...
\end{equation}
We need to calculate each term separately, but if we calculate $\partial_{\kappa_i}\!\!\!{^{*}\!\mathcal{V}_i}$ first, then we can find that:
\begin{equation}
\renewcommand{\arraystretch}{1.5}
\setlength{\arraycolsep}{1.5pt}
\begin{array}{r l}
\partial_{\kappa_i}\!\!\!{^{*}\!\mathcal{V}_i}&=^{*}\!\!\!(\partial_{\kappa_i}\!\mathcal{V}_i)\\
&=^{*}\!\!\!(\partial_{\kappa_i}
\left[\begin{smallmatrix}
l_i\kappa_i R(\varphi_i) \omega\\
l_iR(\varphi_i)  ^{*}\!\hat{q} \omega
\end{smallmatrix}\right])\\
&=
^{*}\!\!\!\left[\begin{smallmatrix}
l_i R(\varphi_i) \omega\\
\mathbf{0}
\end{smallmatrix}\right]\\
&=\left[\begin{smallmatrix}
l_i {}^{*}\!(R(\varphi_i) \omega)&\mathbf{0}\\
\mathbf{0}^\top&0
\end{smallmatrix}\right]
\end{array}
\end{equation}
and the commutator:
\begin{equation}
\renewcommand{\arraystretch}{1.5}
\setlength{\arraycolsep}{1.5pt}
\begin{array}{r l}
[ {}^{*}\!\mathcal{V}_i,\partial_{\kappa_i}\!\!\!{^{*}\!\mathcal{V}_i}]&=\left[\left[
\begin{smallmatrix}
 l_i\kappa_i {}^{*}\!(R(\varphi_i) \omega)&
l_iR(\varphi_i)  ^{*}\!\hat{q} \omega\\
\mathbf{0}^\top&0
\end{smallmatrix}\right],
\left[\begin{smallmatrix}
l_i {}^{*}\!(R(\varphi_i) \omega)&\mathbf{0}\\
\mathbf{0}^\top&0
\end{smallmatrix}\right]\right]\\
&=\left[
\begin{smallmatrix}
 \mathbf{0}&l_i^2\kappa_i R(\varphi_i) {}^{*}\!(\omega)^2 \hat{q}
\\
\mathbf{0}^\top&0
\end{smallmatrix}\right]
\end{array}
\end{equation}
This allows to find that,
\begin{equation}
\renewcommand{\arraystretch}{1.5}
\setlength{\arraycolsep}{1.5pt}
\begin{matrix}
[ {}^{*}\!\mathcal{V}_i,\partial_{\kappa_i}\!\!\!{^{*}\!\mathcal{V}_i}] {}^{*}\!\mathcal{V}_i=\mathbf{0}
,\text{ and}\\
 {}^{*}\!\mathcal{V}_i^n[ {}^{*}\!\mathcal{V}_i,\partial_{\kappa_i}\!\!\!{^{*}\!\mathcal{V}_i}]=\left[
\begin{smallmatrix}
 \mathbf{0}&l_i^{n+2}\kappa_i^{n+1} R(\varphi_i) {}^{*}\!(\omega)^{n+2} \hat{q}
\\
\mathbf{0}^\top&0
\end{smallmatrix}\right]
\end{matrix}
\end{equation}
therefore:
\begin{equation}\label{eq:vk}
\renewcommand{\arraystretch}{1.5}
\setlength{\arraycolsep}{1.5pt}
\begin{array}{r l}
^{*}\!\mathcal{V}^\kappa_i&
=\left[
\begin{smallmatrix}
 l_i {}^{*}\!(R(\varphi_i) \omega)&\quad\kappa_i^{-1}R(\varphi_i)\sum_{n=0}^{\infty}\frac{1}{(n+2)!}l_i^{n+2}\kappa_i^{n+2}  {}^{*}\!(\omega)^{n+2} \hat{q}
\\
\mathbf{0}^\top&0
\end{smallmatrix}\right]\\
&=\left[
\begin{smallmatrix}
 l_i {}^{*}\!(R(\varphi_i) \omega)&\quad\kappa_i^{-1}R(\varphi_i)(e^{l_i\kappa_i\!{}^{*}\!\omega}-l_i\kappa_i^{*}\!\omega-I)\hat{q}
\\
\mathbf{0}^\top&0
\end{smallmatrix}\right]\\
\mathcal{V}^\kappa_i&=\left[
\begin{smallmatrix}
 l_i(R(\varphi_i) \omega)\\
 \kappa_i^{-1}R(\varphi_i)(e^{l_i\kappa_i\!{}^{*}\!\omega}-l_i\kappa_i^{*}\!\omega-I)\hat{q}
\end{smallmatrix}\right] \text{(  for } \kappa_i \neq 0 \text{ )}
\end{array}
\end{equation}
For $^{*}\!\mathcal{V}^\varphi_i$, we choose to use the Hausdorff formula for the twist \cite{Selig2010}:
\begin{equation}\label{eq:vp}
\renewcommand{\arraystretch}{1.5}
\setlength{\arraycolsep}{1.5pt}
\begin{array}{r l}
\mathcal{V}^\varphi_i&=\sum_{n=0}^\infty\frac{1}{(k+1)!}\boldsymbol{ad}^n ({}^*\!\mathcal{V}_i)\partial_{\varphi_i}\!\mathcal{V}_i\\
&=\quad [I+\frac{4-\kappa_il_i\text{sin}(\kappa_il_i)-4\text{cos}(\kappa_i l_i)}{2\kappa_i^2l_i^2}\boldsymbol{ad}({}^*\!\mathcal{V}_i)\\
&\quad +
\frac{4\kappa_i l_i-5\text{sin}(\kappa_il_i)+\kappa_i l_i\text{cos}(\kappa_i l_i)}{2\kappa_i^3l_i^3}\boldsymbol{ad}^2({}^*\!\mathcal{V}_i)\\
&\quad +
\frac{2-\kappa_il_i\text{sin}(\kappa_il_i)-2\text{cos}(\kappa_i l_i)}{2\kappa_i^4l_i^4}\boldsymbol{ad}^3({}^*\!\mathcal{V}_i)\\
&\quad +
\frac{2\kappa_i l_i-3\text{sin}(\kappa_il_i)+\kappa_i l_i\text{cos}(\kappa_i l_i)}{2\kappa_i^5l_i^5}\boldsymbol{ad}^4({}^*\!\mathcal{V}_i)]\partial_{\varphi_i}\!\mathcal{V}_i\\
&\quad\text{(  for } \kappa_i \neq 0 \text{ )}\\
\end{array}
\end{equation}
For the case \(\kappa_i = 0\), we can calculate the limitation of (\ref{eq:vk}) and (\ref{eq:vk}) when \(\kappa_i \rightarrow 0\), the results are:
\begin{equation}
\mathcal{V}^\kappa_i=\left[
\begin{smallmatrix}
 l_i(R(\varphi_i) \omega)\\
\boldsymbol{0}
\end{smallmatrix}\right] \text{\quad( } \kappa_i = 0 \text{ )}
\end{equation}
and
\begin{equation}
\begin{aligned}
    \mathcal{V}^\varphi_i =& I + \frac{1}{2}\boldsymbol{ad} ({}^*\!\mathcal{V}_i)+\frac{1}{6}\boldsymbol{ad}^2 ({}^*\!\mathcal{V}_i)\\
&+\frac{1}{24}\boldsymbol{ad}^3 ({}^*\!\mathcal{V}_i)+\frac{1}{120}\boldsymbol{ad}^4 ({}^*\!\mathcal{V}_i) \text{\quad( } \kappa_i = 0 \text{ )}
\end{aligned}
\end{equation}
For \(\mathcal{V}_i^l\), we have:
\begin{equation}
[^{*}\!\mathcal{V}_i,\partial_{l_i} \!\!{}^{*}\!\mathcal{V}_i] = [\frac{{}^{*}\!\mathcal{V}_i}{l_i},{}^{*}\!\mathcal{V}_i] = \boldsymbol{0}
\end{equation}
Therefore:
\begin{equation}
    ^{*}\!\mathcal{V}^l_i=\partial_{l_i}\!\!{^{*}\!\mathcal{V}_i}\text{ and }\mathcal{V}^l_i=\partial_{l_i}\!\mathcal{V}_i
\end{equation}

\addtolength{\textheight}{-2cm}   








\bibliographystyle{ieeetr}
\bibliography{bibbib}

\end{document}